\documentclass[pdflatex,sn-mathphys]{sn-jnl}

\jyear{2023}%

\theoremstyle{thmstyleone}%
\theoremstyle{thmstyletwo}%

\theoremstyle{thmstylethree}%

\usepackage{soul}

\makeatletter
\newcommand\avsuminner[2]{%
  {\sbox0{$\d@th#1\sum$}%
   \vphantom{\usebox0}%
   \ooalign{%
     \hidewidth
     \smash{\vrule height\dimexpr\ht0+1pt\relax depth\dimexpr\dp0+1pt\relax}%
     \hidewidth\cr
     $\d@th#1\sum$\cr
   }%
  }%
}
\makeatother

\begin{document}

\title[A longitudinal multi-modal dataset for dementia monitoring and diagnosis]{A longitudinal multi-modal dataset for dementia monitoring and diagnosis}


\author*[1]{\fnm{Dimitris} \sur{Gkoumas}}\email{d.gkoumas@qmul.ac.uk}
\author[2]{\fnm{Bo} \sur{Wang}}\email{bwang29@mgh.harvard.edu}
\author[1,3]{\fnm{Adam} \sur{Tsakalidis}}\email{a.tsakalidis@qmul.ac.uk}
\author[3,4]{\fnm{Maria} \sur{Wolters}}\email{maria.wolters@ed.ac.uk}
\author[1,3,5]{\fnm{Matthew} \sur{Purver}}\email{m.purver@qmul.ac.uk}
\author[1]{\fnm{Arkaitz} \sur{Zubiaga}}\email{a.zubiaga@qmul.ac.uk}
\author*[1,3]{\fnm{Maria} \sur{Liakata}}\email{m.liakata@qmul.ac.uk}

\affil*[1]{\orgdiv{School of Electronic Engineering and Computer Science}, \orgname{Queen Mary University of London},   \country{UK}}

\affil*[2]{\orgdiv{Center for Precision Psychiatry}, \orgname{ Massachusetts General Hospital, Boston},   \country{USA}}

\affil*[3]{\orgdiv{The Alan Turing Institute, London},   \country{UK}}

\affil*[4]{\orgdiv{School of informatics}, \orgname{University of Edinburgh},   \country{UK}}

\affil*[5]{\orgdiv{Department of Knowledge Technologies}, \orgname{Jožef Stefan Institute},   \country{Slovenia}}


\abstract{Dementia affects cognitive functions of adults, including memory, language, and behaviour. Standard diagnostic biomarkers such as MRI are costly, whilst neuropsychological tests suffer from sensitivity issues in detecting dementia onset. The analysis of speech and language has emerged as a promising and non-intrusive technology to diagnose and monitor dementia. Currently, most work in this direction ignores the multi-modal nature of human communication and interactive aspects of everyday conversational interaction. Moreover, most studies ignore changes in cognitive status over time due to the lack of consistent longitudinal data. Here we introduce a novel fine-grained longitudinal multi-modal corpus collected in a natural setting from healthy controls and people with dementia over two phases, each spanning 28 sessions. The corpus consists of spoken conversations, a subset of which are transcribed, as well as typed and written thoughts and associated extra-linguistic information such as pen strokes and keystrokes. We present the data collection process and describe the corpus in detail. Furthermore, we establish baselines for capturing longitudinal changes in language across different modalities for two cohorts, healthy controls and people with dementia, outlining future research directions enabled by the corpus.}

\keywords{Longitudinal multi-modal dementia corpus, computational linguistics, longitudinal dementia monitoring}


\maketitle

\section{Introduction}
\label{sec:introduction}
Over 50 million people in the world have dementia, a syndrome involving deterioration in memory and cognitive abilities, with an annual increase of 10 million \cite{world2019risk}. Earlier diagnosis can improve patients' quality of life by enabling better planning and medical interventions at the most effective and appropriate stage~\cite{prince2011benefits, prince2014dementia,alzheimer20162016}--especially if diagnosis occurs before clinical symptoms onset in~\cite{ritchie2016clinical, mortamais2017detecting}. 

Standard diagnostic biomarkers such as MRI, PET scans and cerebrospinal fluids are intrusive and expensive, and therefore unsuitable for early diagnosis. The other main family of diagnosis methods are cognitive tests such as the ADAS-Cog \cite{rosen1984new}, MMSE \cite{folstein1975mini} and ACE-III \cite{noone2015addenbrooke}, widely used in clinical studies. As well as suffering from sensitivity issues \cite{Schneider2009CurrentAD} and lagging behind biomarkers in their ability to detect dementia onset \cite{Jack2013TrackingPP}, they have several other caveats: they are usually administered manually; they are pencil-and-paper type tests requiring an expert, and therefore applied only after initial referral to a doctor; they are unsuited for testing across large populations or at home. For early diagnosis, pre-screening and condition monitoring we need methods that can be automated, complementary to biomarkers but easily observable within everyday life without intrusion.

\paragraph{Language datasets for dementia:} The automatic analysis of patients' spontaneous speech and language is a promising non-invasive, inexpensive approach to screening and monitoring dementia progression, as speech and language impairments caused by dementia can occur early in the course of the disease~\cite{Fraser2015LinguisticFI,Knig2018FullyAS}. To allow progress in this field researchers have been working with a number of different datasets. 
Table~\ref{tbl:dataset_review} provides an overview of the most widely used datasets (along with a new datasets proposed by us) in terms of different aspects such as the population, the amount of data, the modality, the nature of the tasks and ability to elicit linguistic information, the duration of the elicited tasks, and the longitudinal aspect (if any). Most consist of speech obtained in a clinical setting; some include either partially \cite{Hansebo2002CarersIW,Weiner2016DetectionOI} or fully \cite{Luz2020AlzheimersDR} transcribed text  or text extracted through Automatic Speech Recognition (ASR) \cite{Luz2021DetectingCD}. 

\paragraph{Language based tasks:}The majority of research has focused on distinguishing people with Alzheimer’s Disease (AD) from cognitively normal controls, a.k.a., the \emph{AD classification} task. The Pitt~\cite{Becker1994TheNH}, ADReSS~\cite{Luz2020AlzheimersDR}, and $\textup{ADReSSo}$~\cite{Luz2021DetectingCD} datasets have been widely used for addressing this task; Among them Pitt is the largest dataset. The datasets include speech recordings~\cite{Luz2021DetectingCD} or speech and transcripts of verbal descriptions~\cite{Becker1994TheNH,Luz2020AlzheimersDR} obtained by asking subjects to describe either general-purpose pictures (e.g., pictures showing animals) or the Cookie Theft picture (CTP) from the Boston Diagnostic Aphasia Examination~\cite{Goodglass2013BostonDA}. Unlike the Pitt Corpus~\cite{Becker1994TheNH}, which contains annual data for the same person up to five times, ADReSS~\cite{Luz2020AlzheimersDR} and $\textup{ADReSSo}$~\cite{Luz2021DetectingCD} include a single speech sample per participant. ADReSSo also contains occasional interactions between instructors and participants.

The inference of patients' cognitive scores, a.k.a., the \emph{score regression} task, has seen less attention. Several studies have extracted different linguistic and acoustic features to predict cognitive scores using the Pitt dataset~\cite{Becker1994TheNH}. In \cite{Luz2020AlzheimersDR,Luz2021DetectingCD}, authors use acoustic features and features extracted from transcripts in a single and bimodal setting for predicting cognitive scores, showing the effectiveness of simple modality fusion.

\indent The task of predicting changes in cognitive status per individual over time, a.k.a., \emph{disease progression}, has received even less attention due to the lack of consistent longitudinal datasets. The Pitt~\cite{Becker1994TheNH} and  Carolinas~\cite{Pope2011FindingAB} datasets are the largest longitudinal datasets currently available for studying the language of individuals with dementia. An important limitation of the Pitt corpus is that the longitudinal aspect is limited, spanning up to 5 sessions maximum per individual with most participants having two narratives only. On the other hand, in the Carolinas dataset, healthy controls have up to two interviews over the longitudinal study while people with dementia have between 1 and 9 sessions. 
Preliminary work has addressed disease progression as a classification task~\cite{Weiner2016DetectionOI,Clark2016NovelVF,Luz2021DetectingCD} on the basis of longitudinal assessments in ADReSSo spanning two years from dementia onset. However, spontaneous speech is only collected once.

An important limitation of existing dementia datasets is that their longitudinal aspect is limited to data snapshots at either a fixed time~\cite{Luz2020AlzheimersDR,Luz2021DetectingCD} or a few points in time~\cite{Becker1994TheNH,Pope2011FindingAB}. Moreover, the language elicitation tasks are biased towards particular genres or domains via lab-based tasks, such as the description of a particular set of images~\cite{Becker1994TheNH,Luz2020AlzheimersDR,Luz2021DetectingCD}. Some studies elicit speech from more natural spontaneous conversations~\cite{Pope2011FindingAB,Weiner2016DetectionOI}. Yet, the tasks are restricted to particular topics, well known to participants and subject to learning effects~\cite{Goldberg2015PracticeED}. Due to the above limitations in dementia datasets, existing work in language and speech processing for discriminating across cohorts of healthy controls and people with dementia ignore changes in language and how these relate to changes in cognition over time. 


\begin{sidewaystable}
\sidewaystablefn%
\begin{center}
\begin{minipage}{\textheight}
\caption{Overview of the most widely used datasets and comparative benefits of our new proposed multimodal longitudinal dataset. Elicitation = Data elicitation task. Duration: Average speech duration of the elicited task in minutes. CTP = Cookie Theft Picture. Dem=Participants with Dementia. Ctrl=Participants with no dementia diagnosis. AD Class= Alzheimer’s Dementia classification. Score Reg=Score Regression.}
\label{tbl:dataset_review}
\resizebox{\textwidth}{!}{%
\begin{tabular}{p{3cm}ccccp{3cm}cp{3.5cm}}
 \toprule
\textbf{Dataset} & \textbf{Population} & \textbf{Samples} & \textbf{Modality} & \textbf{Longitudinal} & \textbf{Elicitation} & \textbf{Duration} & \textbf{Task} \\
 \hline
DementiaBank \cite{Becker1994TheNH}& 196 Dem vs.\ 98 Ctrl & 255 Dem vs.\ 244 Ctrl  & Audio\textbf{$\dagger$} & 1-5 sessions & CTP Description & 1.90 & AD Class, Score Reg\\
\hline
Carolinas\cite{Pope2011FindingAB} &125 Dem vs 125 Ctrl &400  Dem Vs 250 Ctrl & Text$+$Audio$+$Video\textbf{$\star$} & 1-9 sessions & Health and Well-being Conversations & 76.14  & AD Class, Detection of Confusion\\
\hline
ADReSS\cite{Luz2020AlzheimersDR} & 78 Dem vs.\ 78 Ctrl & 78 Dem vs.\ 78 Ctrl & Text$+$Audio\textbf{$\dagger$} & No  & CTP Description & 1.20 & AD Class, Score Reg\\
\hline
$\textup{ADReSS}_{o}$(1) \cite{Luz2021DetectingCD}  & 32 Dem & 105 & Audio\textbf{$\ddag$} & No & Picture Description & 2.17 & Disease Progression\\ \hline
$\textup{ADReSS}_{o}$(2) \cite{Luz2021DetectingCD} & 36 Dem vs.\ 35 Ctrl &  237 & Audio\textbf{$\ddag$} & No & Picture Description & 2.21 &  AD Class, Score Reg\\ \hline
Carers' interactions with dementia patients \cite{Hansebo2002CarersIW} & 14 Dem & 14 Dem & Text$+$Video\textbf{$\ddag$}& No & Video-recorded Interactions & - & Qualitative Analysis \\ \hline
ILSE \cite{Weiner2016SpeechBasedDO} & 23 Dem & 112 hours recording& Text$+$Audio$\star$ & 1-4 sessions &  Recorded Interactions & - & Disease Progression \\
\hline
Verbal fluency and brain imaging scores \cite{Clark2016NovelVF} & 107 Dem vs.\ 51 Ctrl& - & Text+MRI scores\textbf{$\dagger$}& 1-4 sessions & Recorded Interactions & - & Disease Progression\\ \hline \hline
Our data collection & 14 Dem vs.\ 8 Ctrl & 408 Dem vs.\ 408 Ctrl  & Audio+Text+Keyboard+Pen & 56 sessions &  Reminiscence Materials & 12.25 & Longitudinal Language Changes\\
 \bottomrule
\end{tabular}}
 \footnotetext{\textbf{$\star$} \footnotesize{Conversations, \textbf{$\dagger$} Monologue speech}, \textbf{$\ddag$} Occasional interactions between clinicians and participants.}
\end{minipage}
\end{center}
\end{sidewaystable}
Here we address the above limitations and make the following contributions as follows:  

\begin{itemize}

\item We introduce a novel multi-modal dementia corpus of rich longitudinal natural conversations collected over 2 phases, each spanning 28 sessions. This was obtained from people with various forms of dementia and healthy controls on the basis of reminiscence material and in a non-clinical setting. The corpus contains speech, transcriptions, and written language (i.e., pen and keyboard modalities). To the best of our knowledge, this is the first multi-modal longitudinal dataset with this range of modalities and covering such fine-grained longitudinal spans. We present the data collection process and the dataset itself in detail. 

\item We establish longitudinal tasks and baselines across different modalities and investigate language changes across the cohorts of health controls and people with dementia over time. 

\item We conduct a set of experiments that
show significant discrimination between healthy controls and people with dementia across all modalities. Our tasks and baselines pave the way on future directions enabled by our new dataset.

\end{itemize}

\section{Related work}
\paragraph{Linguistic manifestation of dementia:} Dementia is often associated with reduction in vocabulary size, syntactic complexity and information content \citep{Maxim1994LanguageOT,Croisile1996ComparativeSO}, as well as loss of coherence, both temporal (construction of logical time sequences) and thematic  (continuity of topic) \citep{Ellis1996CoherencePI}. 
When conducting dialogue, adults with dementia show less coherence and cohesion and more disruptive topic shifts and empty phrases \citep{Dijkstra2004ConversationalCD}, more topically irrelevant utterances \citep{StPierre2005LackOC}, and characteristic ways of responding to questions \citep{Elsey2015TowardsDC}. Dementia has also been associated with apathy \citep{Nobis2018ApathyIA}, emotional dysregulation and mood swings, even in people with mild to moderate dementia \citep{Petry1989PersonalityAI}; such emotional aspects are known to surface and are detectable in language \citep[e.g.][]{Purver2012ExperimentingWD}. 

\paragraph{Language tests:}
Interview-based tests have therefore been developed that assess the linguistic ability for diagnosis and disease progression \citep{Taler2008LanguagePI,Tarawneh2012TheCP}. However these tests are vulnerable to practice effects~\cite{Goldberg2015PracticeED}, and are not applicable to everyday spontaneous speech. Promisingly, \cite{ForbesMcKay2013ProfilingSS,ForbesMcKay2014ChartingTD} showed that linguistic characteristics of spontaneous speech and writing can reliably discriminate healthy older controls from mild-moderate AD patients, and track aspects of decline over time; however, longitudinal findings were limited by infrequent (6-monthly) data collection and their results relied on manual linguistic analysis.

\paragraph{NLP for dementia:}
More recent work has used NLP approaches for dementia detection by analysing aspects of language such as lexical, grammatical, and semantic features \cite{Ahmed2013ConnectedSA,Orimaye2017PredictingPA,Kav2018SeverityOA}, showing that people with dementia produce less lexical and semantic context and lower syntactic complexity compared to healthy controls. Lack of fluency through the study of paralinguistic features has also been shown to be indicative of people with dementia \cite{Ipia2013OnTS}. The semantics and pragmatics of language appear to be affected by dementia throughout the entire span of the disease, more so than syntax \cite{Bayles1982ThePO}. In particular, people with dementia talk more slowly with longer pauses \cite{Gayraud2011SyntacticAL,Ipia2013OnTS,Pistono2019WhatHW}.

Recent work has also investigated manually engineered acoustic features to recognize AD from spontaneous speech \cite{Luz2020AlzheimersDR,Luz2021DetectingCD} while other work exploited non-linguistic features to distinguish people with AD from healthy controls \cite{Nasreen2021DetectingAD}. Neural models such as LSTM and CNN\cite{Karlekar2018DetectingLC} or pre-trained language models \cite{Yuan2020DisfluenciesAF}, have been used to analyze disfluency characteristics, such as filled pauses. Researchers have used neural approaches to extract either acoustic features  \cite{Pan2020AcousticFE,Pan2021UsingTO} or linguistic information \cite{Zhu2021WavBERTES} directly from the speech signal for dementia detection. 

\paragraph{Longitudinal language changes:} Existing work has focused on distinguishing people with dementia from healthy controls without considering language changes over time. Moreover, where present, longitudinal data in current datasets are sparse. For example, in the Carolinas Corpus~\cite{Pope2011FindingAB}, the largest available longitudinal dataset, subjects have up to 9 speech records across the longitudinal study. However, only 8 people (3 dementia and 5 controls) have 9 speech records across the entire collection. Our newly introduced corpus consists of 22 people, where each subject was asked to record 28 sessions of 15 mins of speech for each of the two study phases, as well as provide written logs (See Table~\ref{tbl:dataset_review} for the comparative benefits of our dataset). Additionally, ours is the first study to investigate longitudinal language changes across modalities and how these manifest in the dementia and control cohorts.

\section{Collecting the longitudinal multimodal corpus}
\label{sec:demetia_study}
\noindent Our goal has been to collect a longitudinal multi-modal corpus including both spontaneous speech and writing, as well as extra-linguistic information associated with language production, and to focus on interactions occurring in a non-clinical setting. There are three important novel aspects in the corpus design: I) Conversations and written thoughts as well as associated paralinguistic information are obtained on the basis of  \emph{reminiscence} material, specifically images from past decades on topics of general interest. Reminiscence is a meaningful and useful activity for people with and without dementia that can improve cognition, mood, and quality of life~\cite{Pinquart2012EffectsOR,Gonzalez2015ReminiscenceAD}. II) The corpus contains daily data over two phases each spanning around 4 weeks or 28 sessions. III) The corpus is collected in the participants' own environment using a custom-built tablet application.  


\subsection{Corpus collection process}
According to our protocol our corpus is collected in separate phases each lasting four weeks or 28 sessions, where phases are 14 weeks apart. In practice due to unforeseen delays the period between phases has been longer than 14 weeks (see section~\ref{sec:limitations}).  
Participants are paired with a carer and asked to record daily sessions during each study phase, alternating between (a) 15 mins of conversation with their carers, and (b) typed or hand-written thoughts using a stylus pen.

Both spoken and written language are elicited using reminiscence material, i.e., images from the past, created by the dementia communication specialists Many Happy Returns\footnote{Now called Real communications https://realcommunicationworks.com}. Images are presented using a bespoke Android Tablet application which records spoken and written data and sends it to a secure remote server for storage. The application was designed, developed, and tested together with our commercial clinical partner Clinvivo\footnote{https://www.clinvivo.com}, in consultation with a stakeholder group from the Alzheimer's Society. The application allows recording three modalities: \emph{speech}, \emph{typed text (keyboard)}, and \emph{hand-written text (pen)}. For the latter two paralinguistic information such as key strokes and pen strokes, pen pressure and deletions are recorded respectively. 

At the start of each 4-week phase, participants are given a tablet running the purpose built application which contains reminiscence material and allows recording language in the various modalities. 
A Mini Mental State Examination (MMSE) \cite{folstein1975mini} and an Addenbrooke's Cognitive Examination-III (ACE-III) \cite{noone2015addenbrooke} are administered by a suitably qualified person at the start and end of each phase respectively as cognitive impairment benchmarks. 
While these tests provide only minimal cognitive data, they allow us to assess cognitive change at the comparison points needed to analyse the rich linguistic information collected. 

The investigators monitor the submission of data via a remote server. If data have not been submitted for 48 hours, a research team member contacts the participant and carer to ensure they are not having difficulties. If a participant is unable to record data for longer than a week, the research team considers their withdrawal from the study. If it is deemed that a participant has lost capacity to consent, they are withdrawn from further data collection. Subject to consultation with the participants' carers, any data collected for that participant up to that point are used in the analysis. 

\subsection{Reminiscence Material \& Tablet application}
\begin{figure}[hbt!]
\centering
\includegraphics[width=.65\textwidth]{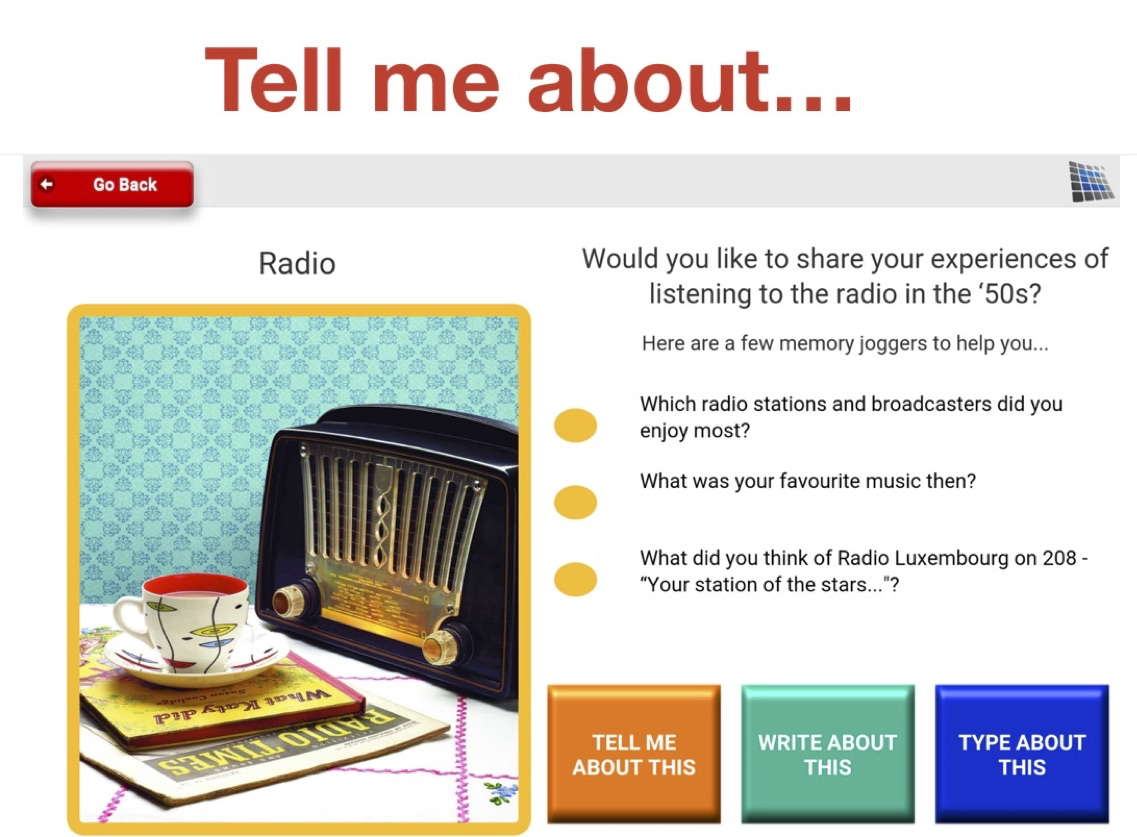}
\caption{Screenshot of Tablet application once topic is chosen (here ``Radio'').}
\label{fig:application}
\end{figure}

The bespoke tablet application shows a participant four images every day, each representing a topic of general interest from the 50s, 60s and 70s. Each 
image/topic is accompanied by 
three questions to help initiate a conversation or thought process and provide memory ``joggers''. The participant chooses one topic out of these four 
as well as the mode of interaction (recording a conversation, typing or writing thoughts). The four images are pseudo-randomly selected from a pool of images
available in each of the 4-week phases. Figure \ref{fig:application} illustrates the table application once a subject has chosen a particular topic (here ``Radio'').

The image material and corresponding questions were developed by an organisation specialising in dementia communication\footnote{\url{https://realcommunicationworks.com/}} and have been used with people in care homes. The 50s and 60s material was adapted for use in the tablet application. The 70s material was created for the purpose of the study. The collected corpus covers a set of 67 images/topics. Phase 1 includes 26 topics from the 50s and Phase 2 includes 41 images from the 60s and 70s. By design topics were meant to be repeated every so often but based on individual's feedback such repetition has been minimised and each phase includes different material. In particular, Phase 1 topics include: Goblin Teasmade, Washday and Smog, Keeping Warm, Household smells, Sundays, Housewives, Budgies, Radio, Television, The cinema and music, The Goons, Weekly children's comics, The Coronation, Holiday Camps, The Modern Era - transport, Endless freedom, Toys and books, School, Knitting, Hair, Teddy boys and teenagers, Bikes, Fashion, Immigration, National Service, Sport.  Phase 2 was augmented with more images from the 70s, which seems to better fit the memory bump of our participant cohort.

\subsection{Participant Recruitment}
\begin{figure}[hbt!]
\centering
\includegraphics[width=.80\textwidth]{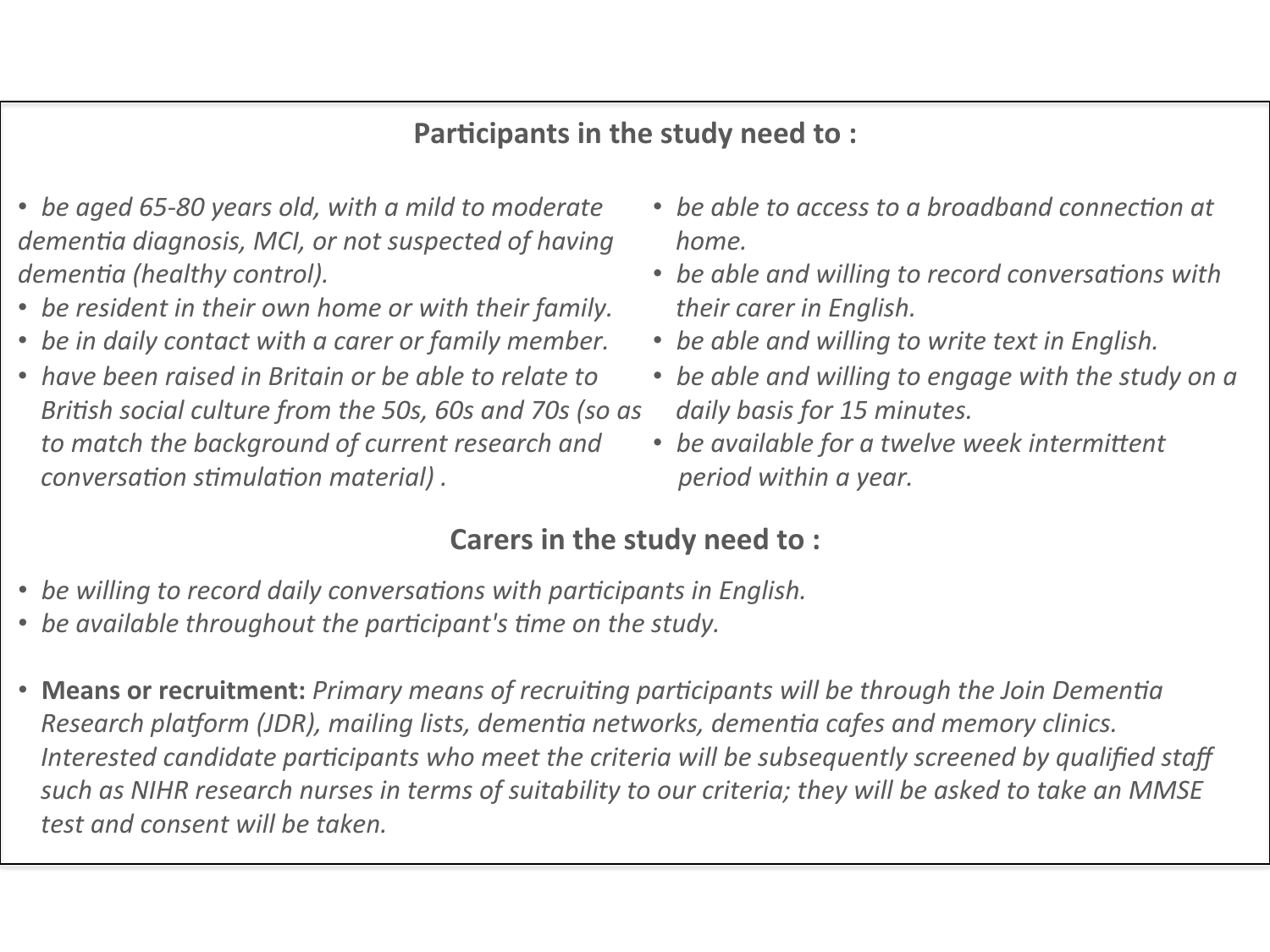}
\caption{Eligibility criteria for study participants and recruitment details.}
\label{fig:eligibility}
\end{figure}

Our target was to recruit a cohort of people living with  dementia or MCI (n=20) and age-matched controls (n=10). These numbers are appropriate for a pilot study. Previous longitudinal research examining language indicators in people with Alzheimer's disease over a year~\cite{ForbesMcKay2013ProfilingSS,ForbesMcKay2014ChartingTD} have recruited similar numbers of participants, with assessments every six months. 
Participant inclusion and exclusion criteria as well as recruitment methods are described in Figure~\ref{fig:eligibility}.

\noindent\textbf{Eligibility:} Participants are aged 65-80 years with mild to moderate dementia, MCI, or are age-matched healthy controls. They are resident in their own home or with family and are in daily contact with a carer or family member. They must have lived in the UK during the 50s-70s so they can relate to the reminiscence material. They must be able to conduct daily conversations and write their thoughts using the provided tablet application.

\noindent\textbf{Recruitment:} Primary means of recruiting participants has been through the Join Dementia Research platform (JDR), mailing lists, dementia networks, dementia cafes and memory clinics. Interested candidate participants are subsequently screened by qualified staff (NIHR research nurses) in terms of providing consent and meeting the study eligibility criteria.

\subsection{Transcription}
A subset of the spoken data (51 sessions from 8 participants spanning several weeks) were transcribed manually by experienced dialogue transcribers, using PRAAT.\footnote{\url{https://www.fon.hum.uva.nl/praat/}} As well as the words spoken, transcripts include significant non-verbal events such as utterance timings, pauses, laughter, crying, yawning, whispering, coughing as well as disfluencies including mis-speaking and reformulation. The transcription convention used was developed on the basis of the CHAT protocol \cite{MacWhinney1992TheCP} and techniques for transcribers \cite{Garrard2011TechniquesFT}. 

\subsection{Community support and dissemination}
We have created a steering committee consisting of six Alzheimer's society volunteers and the founder of Many Happy Returns/Real Comminication Works, an organisation working closely with people living with dementia to help with engagement and dialogue. We have had meetings every several months with the entire group and work closely with a subgroup on the design and usability of the data collection application as well as any concerns that may be faced by participants. This committee has had vital input into the design of the study, the identification of suitable participants as well as dissemination of findings at the end of the study.

The raw data itself cannot be made publicly available as this does not comply with our ethics. Yet, it could be made available to interested parties subject to an NDA agreement. In the future, we also aim to make publicly available pre-trained embeddings for linguistic and audio modalities.

\section{Dataset Description}
\label{sec:datadescription}

\subsection{Participant demographics}
\label{sec:demographics}
We have data from 22 participants (6 females and 4 males with no dementia diagnosis, and 3 females and 9 males with a dementia diagnosis). The average age at the time of recruitment for people with dementia was 70.9 years and for healthy controls 68.9 years. Most participants had at least 10 years of education, with about 25\% having completed a University degree. Conditions represented in the collected corpus include Mild Cognitive Impairment (MCI), Alzheimer's Disease (AD), Vascular Dementia (VD), Frontotemporal Dementia (FD), and Mixed Dementia (MD). 
Overall, the corpus includes 10 healthy controls, 5 people with AD, 2 people with MCI, 2 people with FD, 1 with VD, and 2 with MD (1 AD+VD, 1 AD + Lewy body Dementia), covering 816 sessions and different modalities. Table \ref{tbl:overall_data} summarizes dataset statistics across all modalities.

 For Phase 1, the cohorts include 12 people with dementia or MCI and 10 controls. As managing the longitudinal data collection process is expensive, both in terms of human effort and in terms of the hardware, software and data storage 
 our protocol catered for a small number of participants (20 people with dementia and 10 controls). Similar numbers have been used before in previous longitudinal studies~\cite{ForbesMcKay2013ProfilingSS,ForbesMcKay2014ChartingTD}. While our study protocol targeted recruitment of twice as many people with dementia as healthy controls, since we expect greater homogeneity between controls while dementia can manifest in many different ways, in practice we only managed to recruit 12 people with dementia in the given timeframe.  
For delays primarily external to the study, discussed in section ~\ref{sec:limitations}, only 3 people with dementia and 6 controls were able to complete Phase 2 of the study~\footnote{We discuss dropouts caused by unforeseen delays and how we aim to tackle the relatively small amount of subjects in the Limitations section,~\ref{sec:limitations}}. 

\subsection{Statistical Overview of different Data Modalities}
\begin{table}[ht]
  \centering
  \caption{Overview of the dataset. Dem=Participants with Dementia. Ctrl=Participants with no dementia diagnosis.}
  \label{tbl:overall_data}
  \begin{tabular}{cccccccccc }
    \toprule
    \textbf{Modality} & \multicolumn{3}{c}{\textbf{Participants}}  & \multicolumn{3}{c}{\textbf{\# Sessions}} & \multicolumn{3}{c}{\textbf{\# Topics}} \\ 
    \cmidrule(r){2-4} \cmidrule(r){5-7} \cmidrule(r){8-10}
    & All & Dem & Ctrl &  All & Dem & Ctrl  & All & Dem & Ctrl  \\ \hline
    Speech & 22 & 12 & 10 &  490 & 250 & 240 & 63 & 48 & 60 \\   
    Typed & 17  & 11 & 6 & 271 & 140 & 131 &  65 & 54 & 58\\
   Hand-written & 12 & 7 & 5    & 104 & 59 & 45   & 46 & 36 & 34  \\ \hline
    Overall & 22 & 12 & 10 & 816 & 408 & 408 & 66 & 63 & 65  \\
  \bottomrule
  \end{tabular}
\end{table}

The speech modality is the most popular amounting for 490 sessions by 22 participants, that is 101:26 hours of audio data. The Typed/Key modality follows with 17 participants and 271 sessions while the Hand-written/Pen modality was selected by 12 participants in 104 sessions. In general, controls record a larger amount of sessions compared to participants with dementia (51.1 (STD=13.1) vs 29.1 (STD=11.4)). In each session, participants chose 2.5 topics on average (STD=2.6). In total, the sessions cover 66/67 unique topics.  However, participants with dementia addressed fewer topics in the same number of sessions compared to controls (63 vs 65).

\begin{figure}[hbt!]
\centering
\includegraphics[width=0.7\textwidth]{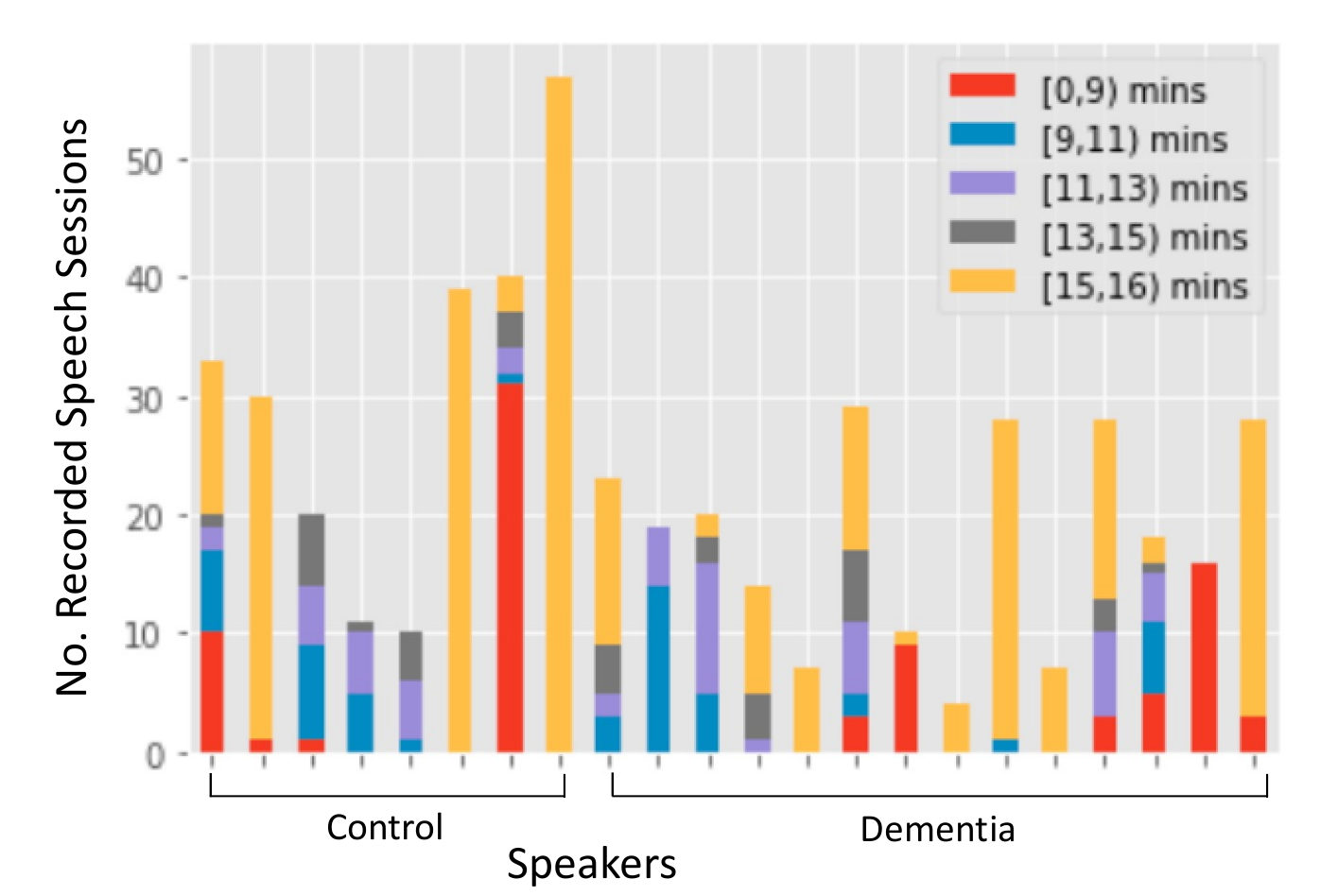}
\caption{Summary of recorded sessions per individual in the two cohorts together with the duration in the speech modality.}
\label{fig:overall_byspeker}
\end{figure}

For speech, the mean duration of sessions is slightly shorter for the dementia group compared to controls (12:11 mins (STD=4:31) vs 12:39 mins (STD=4:07)). Figure \ref{fig:overall_byspeker} summarizes the number of recorded sessions per individual in the two groups together with the session duration in the speech modality. For the majority of sessions conversations last between $15$ and $16$ mins in both groups, although some topics seem harder for both groups, resulting in shorter sessions.  On average, subjects choose the same topic 1.2 (STD=0.5) and 1.3 (STD=0.6) times over the longitudinal study in control and dementia cohorts, correspondingly.  Overall, the duration of individual conversations/sessions is balanced across the two groups. In total, the duration of speech sessions is 50:49 hours for people with dementia and 50:36 hours for controls.

Table \ref{tbl:other_modalities} summarizes statistics of other modalities included in the corpus, i.e., typed and hand-written daily logs, and transcribed daily conversations, along with their corresponding characteristics. For the typed daily logs, healthy controls spend 20.9 minutes writing a log, while the respective length of time to produce a written log for people with dementia is 35.9 minutes. By contrast, the average length of typed characters is 2,647 for healthy controls and 1,752 for people with dementia. We see a similar pattern for the hand-written logs, with the average length of a character sequence produced for this purpose being 529 for healthy controls and 392 for people with dementia. Therefore healthy controls are able to produce more written or typed text within a shorter amount of time. Yet the averaged recorded pen pressure is similar across the two cohorts. 

Part of the spoken conversations (for 8 speakers, 6 people with dementia and 2 controls) has been manually transcribed. Most of the manually transcribed spoken conversations were chosen to be by participants with dementia (79/84 sessions). This allows us in the future to analyse linguistic patterns that characterise  people with dementia and use the corresponding para-linguistic information to fine-tune pre-trained speech-to-text models for automatic speech recognition (ASR) specialising in speech by people with dementia.

\begin{table}[htbp]
\centering 
\caption{Overview of the typed, hand-written, and transcribed conversations in accordance to their particular characteristics. Numbers in parentheses correspond to STD. Chars = Characters} 
\label{tbl:other_modalities}
\resizebox{\columnwidth}{!}{%
\begin{tabular}{lcccccc}
 \toprule
 \multicolumn{1}{c}{} &
 \multicolumn{2}{c}{\textbf{Typed text}} &
 \multicolumn{2}{c}{\textbf{Hand-written text}} & 
 \multicolumn{2}{c}{\textbf{Transcriptions}}\\
 \cmidrule(r){2-3} \cmidrule(r){4-5} \cmidrule(r){6-7}
Group & Duration (mins) &  \# Chars & \# Strokes & \# Chars  & \# People & \# Sessions   \\
 \hline
 Control & 20.9 (23.7)  & 2647 (1639) & 92 (28)  & 529 (409) & 2 & 5 \\
 Dementia & 35.9 (148.3)  & 1752 (1654)  & 87 (21) & 392 (304)  & 6 & 79\\
 \bottomrule
\end{tabular}
}
\end{table}

\section{Longitudinal multimodal language changes across dementia and control cohorts}
\label{sec:task}

\subsection{Task}
Here we showcase the utility of our newly proposed dataset by investigating longitudinal changes in language across different modalities (i.e., speech, transcribed conversation, and typed text) in relation to the two cohorts (i.e., healthy controls and people with dementia). Our goal is to identify subjects' language variations over time. In particular, given a sequence of $N$ sessions $\{S_1, S_2, ..., S_N\}$ over the longitudinal study, we first map each of the sessions to a $d$-dimensional representation $\{S_1^d, S_2^d, ..., S_N^d\}$ such as $d \in \mathbb{I^{+}}$. We then compute the distance $D$ across different sessions over the longitudinal study through cosine similarity for measuring changes in language within subjects. To this effect, we explore two tasks by calculating language changes: a) between adjacent sessions $D(S_t^d,S_{t+1}^d)$ where $t \in N $, called the \textit{consecutive} task, and b) from the beginning of data collection up to time $t$, $D(S_1^d,S_{t}^d)$ where $t \in N $ and $t\gt 1$, called the \textit{non-consecutive} task. For calculating the distance $D$, we consider different statistical functions (i.e., mean, median, std). To the best of our knowledge, this would be the first task to allow such fine-grained multimodal longitudinal analysis as previous work mostly considered modality-specific classification of disease progression at limited fixed time points.

\subsection{Session-level representations}
To obtain session-level representations for both linguistic (transcribed spoken conversations and typed logs) and audio modalities (acoustic aspects of spoken conversations), we first segment language into utterances, where an utterance is defined as an unbroken chain of spoken or written language. We then map each of the utterances into a pre-trained embedding representation. We finally construct session-level representations by averaging the utterance embeddings within sessions.

When working on the linguistic modality, segmentation is performed on an unbroken chain of spoken language for the transcriptions and on punctuation for the typed texts. Each segmented utterance is mapped onto a fixed-size sentence representation~\cite{Reimers2019SentenceBERTSE}. We chose sentence embedding representations as previous work has shown their effectiveness in assessing cognition through language for mental health~\cite{Iter2018AutomaticDO, Voleti2019ARO}.

For the audio modality, we use an end-to-end voice activity detection model~\footnote{\href{https://github.com/pyannote/pyannote-audio-hub}{\tt github.com/pyannote/pyannote-audio-hub}} to perform segmentation on speech. In line with the linguistic modality and as previous work showed the superiority of using neural representations over manual-engineered acoustic features~\cite{Zhu2021WavBERTES}, we map speech segments to pre-trained speech embeddings. Here, we use TRIpLET Loss network (TRILL), which has resulted in a good performance in non-semantic speech tasks including AD classification on DementiaBank~\cite{Shor2020TowardsLA}. We encode moments of silence by applying random initialization.

\subsection{Results}

We calculated the mean, median, and std cosine distance of session-level representations between consecutive and non-consecutive sessions for each speaker individually. We then averaged the obtained scores of speakers across the two cohorts. We chose cosine distance of sentence level representations as it has been shown in previous work to be a strong baseline for tasks in mental health ~\cite{Iter2018AutomaticDO}.

For speech, we noticed that the mean and median cosine distance scores were different across the two cohorts for both the consecutive and non-consecutive tasks (see Table \ref{tbl:results}). However, the distance scores were significantly higher for the dementia group ($p<0.05$) when we calculated changes across non-consecutive sessions. That is changes in speech across sessions were particularly prominent in temporally distant sessions. We also investigated speech variations for people who participated in both phases of the longitudinal study. There are 9 such participants (6 controls and 3 people with dementia). Here, we averaged the session embeddings per participant within a phase and calculated the distance between the two phases. Again, participants in the dementia cohort exhibited substantial speech variations across phases (see Table~\ref{tbl:results_phases}). This justifies further the importance of collecting longitudinal language data for dementia monitoring.

We obtained similar results when conducting experiments with transcriptions and typed texts (see Table~\ref{tbl:results}). Overall, we observed that transcribed speech is most informative in capturing longitudinal language changes across the two cohorts. Yet, speech is more useful when comparing people across phases (see Table~\ref{tbl:results_phases}). In the case of typed text, while distance scores are higher for the dementia cohort, the difference was not statistically significant. 
We assume this is because in planned, non-spontaneous texts, such as written thoughts, the planning going into writing the text makes it more coherent. 
However, the typed and written text modalities convey additional, currently unexplored, extra-linguistic information (number of deletions, pauses between keystrokes), that show corrections of one’s text and these may be better indicators of changes in cognition. In the future, we aim to investigate self-repair tasks~\cite{Rohanian2021BestOB} that are more appropriate for written discourse.

\begin{table}[htbp]
\centering 
\caption{Averaged distance scores between the two cohorts (people with dementia and healthy controls) and across different modalities for the non-consecutive and consecutive tasks. Numbers in bold indicate significant difference across cohorts.} 
\label{tbl:results}
\resizebox{\columnwidth}{!}{%
\begin{tabular}{llcccccc}
 \toprule
 \multicolumn{1}{c}{} &
 \multicolumn{1}{c}{} &
 \multicolumn{3}{c}{\textbf{Non-consecutive}} &
 \multicolumn{3}{c}{\textbf{Consecutive}} \\
 \cmidrule(r){3-5} \cmidrule(r){6-8}
Modality & Group & MEAN& MEDIAN & STD  & MEAN& MEDIAN & STD   \\
 \hline
 Speech & Dementia & \textbf{0.13} & \textbf{0.10}  & {0.11}  & {0.08}  & {0.03} & {0.12} \\
 & Control & \textbf{0.07} & \textbf{0.05} & {0.09}  & {0.06} & {0.01} & {0.12} \\ \hline
 Transcriptions   & Dementia & \textbf{0.16} &  \textbf{0.13}  & 0.11 & 0.11 & 0.08 & 0.11  \\
  & Control & \textbf{0.07}  & \textbf{0.06} & 0.08 & 0.08  & 0.07 & 0.10    \\ \hline
 Typed text \textbf{*}   & Dementia & \textbf{0.011} &   \textbf{0.009} & 0.008 & 0.007  & 0.008  & 0.005  \\
  & Control & \textbf{0.005} & \textbf{0.003} & 0.004 & 0.005 & 0.004 & 0.003   \\ 
 \bottomrule
\end{tabular}
}
\footnotesize{\textbf{*} Results were rounded to the nearest 1000$^{th}$. }
\end{table}

\begin{table}[htbp]
\centering 
\caption{Averaged distance scores between the two phases and across different modalities for subjects participating in both phases of the longitudinal study. Numbers in bold indicate significant difference across cohorts.} 
\label{tbl:results_phases}
\resizebox{\columnwidth}{!}{%
\begin{tabular}{lccccccccc}
 \toprule
 \multicolumn{1}{c}{} &
 \multicolumn{3}{c}{\textbf{Speech}} &
 \multicolumn{3}{c}{\textbf{Transcriptions}} &
 \multicolumn{3}{c}{\textbf{Typed text}}\\
 \cmidrule(r){2-4} \cmidrule(r){5-7} \cmidrule(r){8-10} Group & MEAN& MEDIAN & STD & MEAN& MEDIAN & STD & MEAN& MEDIAN & STD  \\
 \hline
  Dementia & \textbf{0.09} & \textbf{0.04} & \textbf{0.10}  & \textbf{0.12} & \textbf{0.10} & {0.13} & \textbf{0.006}  & \textbf{0.005} & {0.003}\\
  Control & \textbf{0.02} & \textbf{0.01} & \textbf{0.01}  & \textbf{0.06} & \textbf{0.05} & {0.10} & \textbf{0.002} & \textbf{0.001} & {0.001}\\ 
 
 \bottomrule
\end{tabular}
}
\end{table}

\section{Conclusion}
We introduce a novel fine-grained longitudinal multi-modal corpus containing data from healthy controls and people with dementia. The dataset covers audio and text, containing spoken and transcribed conversations, written and typed logs as well as associated extra-linguistic information such as pen and keystrokes. Conversations and written thoughts are elicited in a natural setting, in the participants own environment, triggered by reminiscence material. Specifically, people can record their thoughts via recorded audio, typed or written text through a bespoke tablet application. We present the data collection process and describe the corpus providing statistical information about the two cohorts across the different modalities collected. We also establish baselines to capture longitudinal language changes in relation to the two cohorts and across the audio and linguistic modalities. A set of initial experiments shows that longitudinal language variations are higher in people with dementia. This effect is even more pronounced across temporally distant sessions. In the future, we aim to investigate tasks that involve language-function variations, such as coherence and disfluency, that are particularly prominent in the progression of dementia.

\section{Limitations}
\label{sec:limitations}
In this work, we introduced a multi-modal longitudinal corpus for monitoring changes in dementia progression. The corpus was collected in a natural setting from healthy controls and people with dementia over two phases, each spanning 28 sessions. Moreover, subjects could choose to hold conversations or write or type their thoughts on a variety of topics from reminiscence material provided by a bespoke tablet application. Despite the novel fine-grained longitudinal multimodal nature of the corpus, an important limitation is the relatively small-scale cohorts in the study. In particular there is only a small number of people who where able to participate in the second phase of the study. This was due to unforeseen disruptions to the study first via the introduction of GDPR regulation in 2018, which required pausing of the study to update software for data collection, and then COVID-19. This meant that in several cases 12 months or longer elapsed between phases 1 and 2 and as a result many of our participants were no longer able to participate, primarily due to a decline in their health or change in their personal circumstances. We aim to address this limitation by expanding the existing corpus with a new data collection spanning three phases within twelve months, by recruiting individuals from a collaborating memory clinic. Nevertheless the existing dataset is the first of its kind and has opened new avenues for research in longitudinal changes in language for people with dementia and across different modalities. 

Indeed, we have introduced baselines to capture longitudinal changes in language across modalities in the two cohorts. In particular, we calculated the distance between adjacent and across non-adjacent sessions when those were mapped to fixed-size representations. A set of initial experiments showed promising results for monitoring dementia using fine-grained multi-modal longitudinal data. However these approaches are limited in capturing various linguistic functions associated with the progression of dementia~\cite{tang2008assessment,klimova2015}. In future work, we aim to use NLP techniques  to characterise the language in terms of features likely to be associated with disease onset and progression and/or be suitable for detecting changes in use over time across all types of conversations, i.e., speech, transcriptions, typed and written thoughts. These will include analysis in terms of lexical, syntax and coherence features already identified in the literature \  \cite{Fraser2015LinguisticFI,Ellis1996CoherencePI}; and in terms of recent approaches which infer vector-based representations of words or speakers (\textit{embeddings}) from observed use and are well suited to tracking changes in language use over time \cite{Hamilton2016CulturalSO,tsakalidis2022identifying}.

\section{Ethical Considerations}
\label{sec:ethical}

The collection of the corpus involves ethical considerations especially as we are working with vulnerable individuals who have dementia. The study has received ethics approval from the NHS Research Ethics Committee (REC) and the Health Research Authority (HRA), with reference number 16/WS/0226. 
Participating individuals as well as their carers consented to permit data collection and analysis for research purposes. User identifying information was kept separate from the language data collected via the bespoke tablet application.

While data was collected anonymously, there are potential ethical concerns with using spoken language and computational approaches for monitoring changes in cognitive status and dementia. One concern is related to privacy and confidentiality, as language data may contain sensitive personal information. Other potential risks involve the misuse of models trained on the data for monitoring changes in cognition, which could be used carelessly or maliciously without considering the impact and social consequences in the broader community. To mitigate such risks, we apply strategies such as running software on authorised servers only, with encrypted data during transfer, anonymization of data prior to analysis. Data is only accessed by authorised individuals and interested parties can only obtain access 
subject to an NDA agreement which carefully states research goals.

For a real-world application, ethical concerns are related to the potential for misdiagnosis or overdiagnosis, which could lead to unnecessary treatment or psychological distress for patients and their families. Additionally, there may be issues related to access and equity, as some individuals may not have access to the necessary technology or resources for speech recognition and monitor through analysis of language. Finally, there may be concerns related to the accuracy and reliability of  technology, as well as the potential for bias in the data or algorithms used for monitoring changes. It is important to consider these ethical concerns when developing and implementing technologies for dementia monitoring and diagnosis.

\section*{Acknowledgements}
This work was supported by a UKRI/EPSRC
Turing AI Fellowship to Maria Liakata (grant
EP/V030302/1), the Alan Turing Institute (grant
EP/N510129/1), and Wellcome Trust MEDEA (grant 213939). Matthew Purver acknowledges financial support from the UK EPSRC via the projects Sodestream (EP/S033564/1) and ARCIDUCA (EP/W001632/1), and from the Slovenian Research Agency grant for research core funding P2-0103.









\bibliography{sn-article}


\end{document}